\documentclass[11pt]{article}

\usepackage[utf8]{inputenc}
\usepackage[T1]{fontenc}
\usepackage{lmodern}
\usepackage[margin=1in]{geometry}

\usepackage{amsmath}
\usepackage{amssymb}
\usepackage{graphicx}
\usepackage{float}
\usepackage{booktabs}
\usepackage{array}
\usepackage{tabularx}
\usepackage{multirow}
\usepackage{enumitem}
\usepackage{caption}
\usepackage{changepage}
\usepackage{xcolor}
\usepackage{cite}
\usepackage[colorlinks=true,linkcolor=blue,citecolor=blue,urlcolor=blue]{hyperref}

\newcommand{\hl}[1]{#1}            
\newcommand{\highlighting}[1]{#1}  
\newcommand{\textls}[2][]{#2}      
\newcommand{\reftitle}[1]{}        
\newcommand{\PublishersNote}[1]{}
\newcommand{\authorcontributions}[1]{\medskip\par\noindent\textbf{Author Contributions:} #1\par}
\newcommand{\funding}[1]{\medskip\par\noindent\textbf{Funding:} #1\par}
\newcommand{\dataavailability}[1]{\medskip\par\noindent\textbf{Data Availability Statement:} #1\par}
\newcommand{\acknowledgments}[1]{\medskip\par\noindent\textbf{Acknowledgments:} #1\par}
\newcommand{\conflictsofinterest}[1]{\medskip\par\noindent\textbf{Conflicts of Interest:} #1\par}
\newcommand{\abbreviations}[2]{\medskip\par\noindent\textbf{#1}\par\nobreak\smallskip #2\par}
\newlength{\extralength}
\newlength{\fulllength}
\newcolumntype{C}{>{\centering\arraybackslash}X}

\title{\bfseries Physics-Informed CNN-LSTM for Street-Scale Urban Flood Prediction:\\[2pt]
Reconciling Aggregate Accuracy and Street-Level Plausibility}

\author{%
\parbox{\textwidth}{\centering
\large Luc D'Costa$^{1,2,\ast}$\quad Yidi Wang$^{2,3}$\quad Jonathan L. Goodall$^{2,3,\dagger}$\quad Rohan Chandra$^{1,2}$\\[8pt]
\normalsize
$^{1}$\,Chandra Robot Autonomy Lab, Department of Computer Science, University of Virginia, Charlottesville, VA 22904, USA\\
$^{2}$\,Link Lab, Multidisciplinary Research Center, University of Virginia, Charlottesville, VA 22904, USA\\
$^{3}$\,Hydroinformatics Research Group, Department of Civil and Environmental Engineering, University of Virginia, Charlottesville, VA 22904, USA\\[6pt]
\footnotesize $^{\ast}$\,Corresponding author: dcosta.luc@gmail.com (L.D.);\quad $^{\dagger}$\,goodall@virginia.edu (J.L.G.)%
}%
}
\date{}

\hypersetup{
  pdftitle={Physics-Informed CNN-LSTM for Street-Scale Urban Flood Prediction},
  pdfauthor={Luc D'Costa, Yidi Wang, Jonathan L. Goodall, Rohan Chandra}
}

\begin{document}
\maketitle

\begin{abstract}
\noindent Deep learning surrogate models trained with mean-squared-error loss produce statistically accurate but physically unconstrained flood predictions: water may flow uphill, appear spontaneously, or smooth over street-level corridors. In this work, a physics-informed training framework is developed for CNN-LSTM models that predict urban flood depths at 15 min intervals over a $128\times128$ spatial grid. Three differentiable penalty terms are embedded directly into the loss function: (i) a gravity loss that penalizes depth increases against the water-surface-elevation gradient, (ii) a continuity loss enforcing local mass conservation with rainfall-adaptive thresholds, and (iii) a topography-aware false-alarm penalty modulated by the topographic wetness index (TWI). The framework is evaluated on the Norfolk, Virginia, flood dataset spanning two major storm events (August 2017 and September 2022) comprising 300 samples, with all variants trained on identical splits and robustness assessed over repeated random splits and leave-one-storm-out tests. A road-proximal evaluation restricted to a TWI-derived street mask quantifies street-level skill. The physics-constrained model achieves near-zero gravity violations ($\sim$10$^{-6}$) and the highest street-channel recall (0.77 $\pm$ 0.09 versus 0.44 $\pm$ 0.10 for the unconstrained baseline), the capability most relevant to downstream traffic routing, and its recall advantage more than doubles on a held-out storm, while a uniform false-alarm variant attains 16\% lower mean absolute error but suppresses street recall to 0.25. The proposed TWI-modulated penalty reconciles this trade-off: it improves upon the uniform variant on every metric measured, recovering 60\% higher street recall at the lowest MAE among all constrained variants and the best street-level F1 score. These results expose a fundamental tension between aggregate pixel-level error metrics and application-specific physical plausibility, and demonstrate that terrain-aware loss modulation offers a principled resolution.
\end{abstract}

\medskip


\section{Introduction}
\label{sec:intro}

Flooding causes considerable damage to people, infrastructure, and economies worldwide, with urban flood events accounting for billions of dollars in losses annually \cite{teng2017flood}. Climate change and rapid urbanization are intensifying both the frequency and severity of flood events, particularly in low-lying coastal cities where compound flooding from storm surge, rainfall, and tidal interactions creates complex inundation patterns \cite{mosavi2018flood}. Real-time flood prediction is essential for emergency management and, increasingly, for intelligent transportation systems that must reroute traffic away from inundated roads before they become impassable.

Physics-based hydrodynamic models, numerical solvers for shallow water equations (SWEs), provide physically consistent flood predictions \cite{bates2010simple,brunner2016hec}. However, their computational cost is prohibitive for real-time applications: a single high-resolution simulation of a storm event can require hours on high-performance computing clusters, far exceeding the sub-minute inference budgets demanded by operational systems. This computational bottleneck has motivated surrogate-modeling approaches that use deep learning to approximate the input--output relationship of physics-based models at a fraction of the cost \cite{guo2021data,bentivoglio2022deep}. \mbox{Hu et al.~\cite{hu2019rapid}} demonstrated that integrated LSTM-ROM frameworks can reduce CPU cost by three orders of magnitude compared to full model simulations while maintaining accuracy.

Recent CNN-LSTM architectures have demonstrated strong performance on spatio-temporal flood forecasting tasks. Wang et al.~\cite{wang2026hybrid} developed a hybrid model combining DeepLabv3-style spatial encoding \cite{chen2018encoder} with LSTM temporal processing \cite{shi2015convolutional} for hyper-resolution flood prediction in Norfolk, Virginia, achieving sub-centimeter MAE on their test set. However, this and similar purely data-driven surrogate models are trained with standard loss functions (MSE, MAE) that are agnostic to the governing laws of fluid dynamics. While such models may achieve low aggregate error, they can produce physically implausible predictions: water accumulating at topographic high points, mass appearing spontaneously between timesteps, or street-level flood corridors being smoothed into broad shallow regions.

For safety-critical downstream applications such as traffic routing, the distinction between statistical accuracy and physical plausibility is consequential. A prediction that is accurate on average but misses the specific road segment where water pools provides false confidence to routing algorithms that may direct vehicles into flooded corridors. Conversely, a model with slightly higher aggregate error that correctly identifies which streets flood provides actionable intelligence for road closure decisions.

In this work, hydrological constraints are embedded directly into the training loss function. Rather than modifying the network architecture, the standard data loss is augmented with differentiable penalty terms encoding three physical principles: (1) gravitational flow---depth increases should not align with increasing water surface elevation; (2) mass conservation---water cannot appear or vanish without a source or sink; and (3) topography-aware spatial continuity---false-alarm penalties are modulated by terrain morphology so that predictions in topographic valleys are penalized less than predictions on ridges.

The layout of this paper is as follows. Section~\ref{sec:related} reviews related work on flood modeling, physics-informed neural networks, and loss function design. Section~\ref{sec:method} describes the methodology, including the base CNN-LSTM architecture and the physics-informed loss components. Section~\ref{sec:experiments} presents the experimental setup and results. Section~\ref{sec:discussion} discusses the findings and their implications. Section~\ref{sec:conclusion} provides conclusions and directions for \mbox{future work.}

\smallskip
The main contributions of this work are:
\begin{itemize}[leftmargin=*,labelsep=5.8mm]
    \item A multi-component physics-informed loss function comprising gravity, continuity, and TWI-modulated false-alarm terms, applicable to any differentiable flood-\mbox{prediction architecture.}
    \item A warm-up scheduling strategy that ramps physics constraints gradually during training to prevent early instability.
    \item An empirical analysis revealing a fundamental tension between aggregate error metrics and application-specific physical plausibility, demonstrated through systematic ablation across five model variants and a generic-physics comparator, with robustness assessed over repeated random splits and leave-one-storm-out tests.
    \item A road-proximal evaluation protocol that quantifies street-level predictive skill by restricting recall, precision, and F1 to a street mask derived from the topographic wetness index, requiring no road-network data.
    \item A Keras~3 (v3.15.0) implementation using a custom \texttt{Model} subclass for physics losses that require access to model inputs during training.
\end{itemize}

\section{Related Work}
\label{sec:related}

\subsection{Physics-Based and Data-Driven Flood Modeling}

Traditional flood modeling solves the shallow water equations or Saint-Venant equations numerically. Established tools such as HEC-RAS \cite{brunner2016hec} and LISFLOOD-FP \cite{bates2010simple} provide physically grounded predictions but at high computational cost, limiting their use in real-time operational settings. Teng et al.~\cite{teng2017flood} provide a comprehensive review of flood inundation modeling methods and their trade-offs.

Data-driven surrogates trained on simulation outputs have emerged as real-time alternatives. Guo et al.~\cite{guo2021data} demonstrated CNN-based flood emulation achieving order-of-magnitude speedups over physics-based models. Bentivoglio et al.~\cite{bentivoglio2022deep} surveyed deep learning methods for flood mapping, identifying spatio-temporal prediction and class imbalance as key open challenges. Hu et al.~\cite{hu2018deep} demonstrated that LSTM networks can simulate rainfall-runoff processes effectively from multi-decadal event records. \mbox{Hu et al.~\cite{hu2019rapid}} proposed an integrated LSTM and reduced-order-model (ROM) framework that achieves spatio-temporal flood prediction with CPU reductions of three orders of magnitude. \mbox{Kao et al.~\cite{kao2020exploring}} explored encoder--decoder LSTM architectures for multi-step-ahead flood forecasting, and Xu et al.~\cite{xu2023deep} demonstrated that Transformer-based transfer learning can improve flood prediction in data-sparse basins by leveraging knowledge from data-rich regions. Le et al.~\cite{le2019application} demonstrated near-perfect Nash--Sutcliffe efficiency in LSTM-based multi-day flood forecasting on the Da River basin, confirming recurrent architectures as reliable surrogates for complex nonlinear flood dynamics.

Wang et al.~\cite{wang2026hybrid} developed the CNN-LSTM model that we build upon, combining DeepLabv3-style spatial encoding with three stacked LSTM layers for temporal rainfall processing, targeting the Norfolk, Virginia study area with 15 min resolution predictions.

\subsection{Physics-Informed Neural Networks}

\textls[-15]{The physics-informed neural network (PINN) paradigm, introduced by \mbox{Raissi et al.~\cite{raissi2019physics}},} embeds partial-differential-equation residuals directly into the loss function, enabling networks to satisfy governing equations without labeled data at collocation points. The broader principle of theory-guided data science, using domain knowledge to constrain data-driven models, has been formalized by Karpatne et al.~\cite{karpatne2017theory} and surveyed extensively by Willard et al.~\cite{willard2022integrating}.

In hydrology, physics constraints have been applied to diverse prediction tasks. \mbox{Read et al.~\cite{read2019process}} combined process-based lake models with deep learning for water-temperature prediction. Hoedt et al.~\cite{hoedt2021mc} introduced MC-LSTM, which architecturally enforces mass conservation for rainfall-runoff modeling by designing gate structures that redistribute rather than create or destroy mass.

Our approach differs from standard PINNs in that we do not enforce PDE residuals point-wise but instead encode higher-level hydrological principles, gravity, mass conservation, topographic consistency, as soft penalty terms. Woo et al.~\cite{woo2025physics} applied this paradigm to urban pluvial flooding, training a CNN-based surrogate on physics-model outputs and validating it against both synthetic and historical events---demonstrating that physics-guided surrogate models can generalize to unseen rainfall events at significant computational speedup. This design choice reflects the surrogate-modeling context: we do not seek to solve the SWE directly but rather to constrain a data-driven model to produce outputs that respect fundamental physical principles.

These families of physics-informed methods are not directly interchangeable, which shapes the experimental comparisons that are meaningful in this study. MC-LSTM~\cite{hoedt2021mc} enforces conservation \hl{architecturally} %
 through mass-redistributing gate structures, but is formulated for lumped (basin-scale, non-spatial) rainfall-runoff sequences and does not transfer directly to dense $128\times128$ gridded depth prediction with a convolutional decoder. Point-wise PINN residuals~\cite{raissi2019physics} require the governing PDE state (velocities as well as depths), which the depth-only surrogate setting does not provide. The approach of \mbox{Woo et al.~\cite{woo2025physics}} is methodologically closest to ours but is demonstrated on a different study area and prediction task, precluding a like-for-like quantitative comparison on the Norfolk dataset. For these reasons, the evaluation in this work isolates the contribution of the proposed loss terms through a controlled ablation on a fixed architecture and dataset (Section~\ref{sec:experiments}), rather than through cross-method benchmarking; extending physics-informed baselines to a common gridded benchmark remains an open community need.

\subsection{Loss Function Design for Spatial Prediction}

Class imbalance is a well-documented challenge in spatial prediction tasks. In flood forecasting, typically $\sim$95\% of pixels represent dry ground, creating a strong incentive for models to predict zero everywhere. Lin et al.~\cite{lin2017focal} introduced focal loss for dense object detection, down-weighting well-classified examples to focus learning on hard cases. Sudre et al.~\cite{sudre2017generalised} proposed generalized Dice loss for highly imbalanced medical-image segmentation. Our approach combines asymmetric weighting for class imbalance with physics-specific penalty terms, representing a hybrid of statistical and physical loss design tailored to the flood-prediction domain. Different neural network architecture design has also been explored in other domains such as traffic forecasting~\cite{chandra2019robusttp, chandra2020forecasting, chandra2019traphic} and affective computing~\cite{bhattacharya2020step, mittal2020m3er, mittal2020emotions, mittal2020emoticon}.

\section{\highlighting{Materials and Methods}} %
\label{sec:method}

\subsection{Problem Formulation}

Given an 11-channel spatial input \hl{tensor} %
 $x \in \mathbb{R}^{128\times128\times11}$ and a rainfall temporal signal, the objective is to predict water depth at four future timesteps:
$\hat{y} = f_\theta(x) \in \mathbb{R}^{128\times128\times4}$, corresponding to $t\!+\!15$, $t\!+\!30$, $t\!+\!45$, and $t\!+\!60$ min. The 11 input channels comprise: a Digital Elevation Model (DEM, channel~0), eight historical water-depth observations (channels~1--8), the most recent observed depth (channel~9), and a rainfall/tidal signal encoded as a $12\times8$ sub-region (channel~10). The DEM encodes terrain elevation at each grid cell, which is central to the physics constraints described below.

\subsection{Base Architecture}

\begin{figure}[H]
      \includegraphics[width=\textwidth]{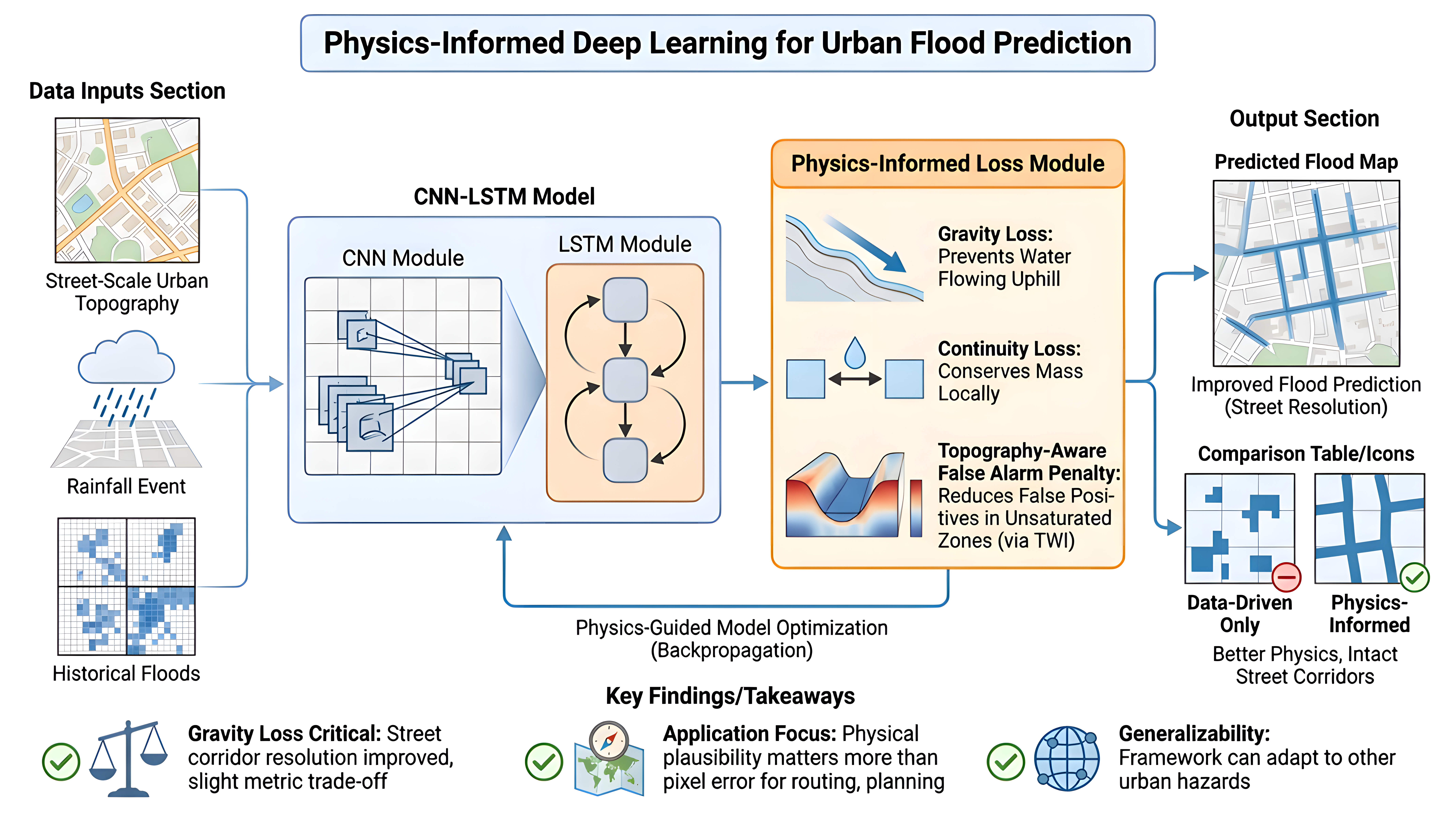}
    \caption{\hl{CNN-LSTM} %
 architecture overview. The spatial encoder (ResNet50 + ASPP) extracts multi-scale spatial features from the 11-channel input. The temporal encoder (3-layer LSTM) processes the rainfall signal. The decoder fuses both streams with skip connections and progressively upsamples to produce $128\times128$ flood-depth predictions at four future timesteps.}
    \label{fig:architecture}
\end{figure}

The CNN-LSTM architecture of Wang et al.~\cite{wang2026hybrid} is adopted, illustrated in Figure~\ref{fig:architecture}. The model comprises three components:

\hl{Spatial encoder.} %
 A \hl{ResNet50 backbone} %
 (pretrained on ImageNet with early layers frozen) extracts spatial features, followed by Atrous Spatial Pyramid Pooling (ASPP) \cite{chen2018encoder} with dilation rates $\{1, 6, 12, 18\}$ for multi-scale feature extraction. A $1\times1$ convolutional adapter maps the 11 input channels to 3 channels for \hl{ResNet} compatibility.

\hl{Temporal encoder.} Three stacked LSTM layers (1024~units each, 20\% dropout) process the rainfall/tidal signal as a sequence of 12~timesteps $\times$ 8~spatial features, producing a temporal context vector.

\hl{Decoder.} ASPP features, skip connections from the encoder's early layers, and the LSTM temporal context are fused and progressively upsampled to $128\times128\times4$ via transposed convolutions with squeeze-and-excitation attention blocks.

  \vspace{-3pt}   

\subsection{Physics-Informed Loss Function}

The total training loss combines a data-fidelity term with physics-based penalty terms:
\begin{equation}
    \mathcal{L}_{\text{total}} = \mathcal{L}_{\text{data}}
    + w_g(t)\,\mathcal{L}_{\text{gravity}}
    + w_c(t)\,\mathcal{L}_{\text{continuity}}
    + \mathcal{L}_{\text{fa}}
    \label{eq:total_loss}
\end{equation}

\noindent where $w_g(t)$ and $w_c(t)$ are epoch-dependent warm-up weights (Section~\ref{sec:warmup}).

\subsubsection{Weighted Mean Squared Error}

To address the $\sim$95\%/5\% dry/wet pixel imbalance, asymmetric weighting is applied:
\begin{equation}
    \mathcal{L}_{\text{data}} = \frac{1}{N}\sum_{i=1}^{N} w_i\,(y_i - \hat{y}_i)^2, \qquad
    w_i = \begin{cases} 10 & \text{if } y_i > 0 \\ 1 & \text{otherwise} \end{cases}
    \label{eq:wmse}
\end{equation}
\noindent \hl{The} %
 $10\times$ wet pixel weighting forces the model to prioritize accuracy on flooded areas rather than the dominant dry majority.

\subsubsection{Gravity Loss (Downhill-Flow Constraint)}

Water flows downhill. Predictions where depth increases align with increasing water surface elevation (WSE$_k$ = DEM + $h_k$) are penalized. For each pair of consecutive timesteps $k$ and $k\!+\!1$, the depth change $\Delta h_k = h_{k+1} - h_k$ is computed. The gravity loss penalizes the component of depth increase that points uphill:
\begin{equation}
    \mathcal{L}_{\text{gravity}} = \frac{1}{4}\sum_{k=0}^{3}
    \mathbb{E}\!\left[
        \operatorname{ReLU}\!\left(
            \nabla(\Delta h_k) \cdot \nabla(\text{WSE}_k)
        \right) \cdot \left(|\Delta h_k| + \epsilon\right)
    \right]
    \label{eq:gravity}
\end{equation}
\noindent \hl{The dot} product $\nabla(\Delta h_k) \cdot \nabla(\text{WSE}_k)$ is positive when depth-change gradients align with WSE gradients (i.e., water increasing uphill). The ReLU activation ensures only uphill flow is penalized. Spatial gradients are computed via finite differences, and $\epsilon = 10^{-6}$ is a small stabilizing constant.

\subsubsection{Continuity Loss (Mass Conservation)}

The continuity loss checks whether depth changes are locally balanced using $5\times5$ average pooling. A dynamic threshold accounts for rainfall as a legitimate water source:
\begin{equation}
    \epsilon_{\text{rain}} = \epsilon_{\text{base}} + \alpha \cdot \overline{r}
    \label{eq:epsilon}
\end{equation}
\noindent where $\overline{r}$ is the mean rainfall intensity and $\alpha$ is a scaling factor. The penalty is asymmetric:
\begin{equation}
    \mathcal{L}_{\text{continuity}} = \frac{1}{4}\sum_{k=0}^{3}\left[
        \mathbb{E}\!\left[\operatorname{ReLU}(\overline{\Delta h}_k - \epsilon_{\text{rain}})^2\right]
        + 0.1\,\mathbb{E}\!\left[\operatorname{ReLU}(-\overline{\Delta h}_k - \epsilon_{\text{rain}})^2\right]
    \right]
    \label{eq:continuity}
\end{equation}
\noindent \hl{Water creation} beyond the rainfall allowance is strongly penalized, while excessive drainage receives a mild penalty ($0.1\times$) since infiltration and subsurface flow are physically expected processes. In all experiments, $\epsilon_{\text{base}} = 0.01$ and $\alpha = 0.5$ (normalized depth units), and $\overline{\Delta h}_k$ denotes the depth change averaged over the $5\times5$ neighborhood.

It should be emphasized that Equation~\eqref{eq:continuity} is a soft, neighborhood-averaged surrogate for mass conservation rather than a discretization of the continuity equation of the shallow water equations: the horizontal flux divergence term $\nabla \cdot q$ is not modeled explicitly, since the surrogate predicts depth fields only and no velocity information is available. Averaging $\Delta h_k$ over a local window instead tests whether depth changes are locally balanced---water leaving one cell should appear in nearby cells---which penalizes spontaneous large-scale creation or destruction of water while remaining agnostic to the direction of flow. The $5\times5$ window was chosen as the smallest neighborhood over which surface water can plausibly redistribute within one 15 min prediction interval at the dataset's grid resolution; substantially smaller windows would penalize legitimate cell-to-cell routing, while substantially larger windows would dilute violations below detectability. This constraint is therefore appropriate for penalizing non-physical water creation in surrogate models, but it should not be interpreted as enforcing exact mass balance, and it does not capture directional momentum effects.

\subsubsection{TWI-Modulated False-Alarm Penalty}\label{sec3.3.4}

A uniform false-alarm penalty (penalizing all predictions on dry ground equally) is topographically blind: it cannot distinguish a narrow road channel, where flooding is physically plausible, from a rooftop, suppressing both. The penalty is modulated using the topographic wetness index (TWI):
\begin{equation}
    \text{TWI} = \ln\!\left(\frac{A}{\tan\beta}\right)
    \label{eq:twi}
\end{equation}
\noindent where $A$ is the upslope contributing area and $\beta$ is local slope. High TWI identifies valleys and street channels; low TWI identifies ridges. The false-alarm penalty becomes
\begin{equation}
    \mathcal{L}_{\text{fa}}^{\text{TWI}} = w_{\text{fa}} \cdot
    \mathbb{E}\!\left[\left(
        1_{\text{dry}} \cdot \operatorname{ReLU}(\hat{y})
        \cdot \underbrace{(1 - 0.85 \cdot \text{TWI}_{\text{norm}})}_{\text{multiplier}}
    \right)^{\!2}\right]
    \label{eq:twi_fa}
\end{equation}
\noindent \hl{Valley pixels} (TWI$_{\text{norm}} \approx 1$) receive a multiplier of 0.15 (85\% penalty reduction), while ridge pixels (TWI$_{\text{norm}} \approx 0$) retain the full penalty. TWI$_{\text{norm}}$ is obtained by min--max normalizing the raw TWI to $[0,1]$ independently for each $128\times128$ input tile (not over the full study-area DEM), so the modulation always spans the full $[0.15, 1]$ range within every training sample. The contributing area $A$ is approximated from the normalized elevation (lower cells accumulate more upslope area), the slope $\tan\beta$ is computed by finite differences with a floor of $0.01$ normalized units to prevent division blow-up on flat terrain, and the TWI is computed from the DEM channel already present in the input, requiring no additional data or architectural modification.

\subsubsection{Warm-Up Scheduling}
\label{sec:warmup}

Physics loss weights are ramped linearly from zero to their target values over the first $T_{\text{warmup}}$ epochs:
\begin{equation}
    w(t) = \min\!\left(1,\;\frac{t+1}{T_{\text{warmup}}}\right) \cdot w_{\text{target}}
    \label{eq:warmup}
\end{equation}
\noindent \hl{Early} in training, the model produces essentially random spatial predictions. Enforcing strict physics constraints on random outputs would dominate the gradient signal and fight the data loss, slowing or preventing convergence. The warm-up allows the model to first learn basic spatial structure from data, then gradually incorporate physical constraints as predictions become meaningful.

\subsection{Implementation Details}

Keras~3 Functional models do not support \texttt{add\_loss()} for losses requiring access to model inputs. A \texttt{FloodModel} subclass of \texttt{keras.Model} is created that overrides \texttt{train\_step()}. During each forward pass, the DEM (channel~0), last observed depth (channel~9), and rainfall signal (channel~10) are extracted from the input tensor, and physics losses are computed on real tensors and added to the total loss before backpropagation. A \texttt{PhysicsLossWarmup} callback updates the epoch counter used by the ramping schedule. All tensor operations use \texttt{keras.ops} for backend-agnostic compatibility.

\section{Results}
\label{sec:experiments}

\subsection{Dataset}

The Norfolk, Virginia, flood dataset from Wang et al.~\cite{wang2026hybrid} is used, comprising high-fidelity hydrodynamic simulation outputs for two major storm events: Hurricane Harvey's remnants (29~August~2017) and a nor'easter (30~September~2022), with exactly 150~samples per event (300 total). Each sample consists of a $128\times128\times11$ input tensor and a \mbox{$128\times128\!\times4$} ground-truth output at 15 min forecast horizons. Data augmentation consists of random horizontal/vertical flips and $90^\circ$ rotations applied to the spatial channels; the rainfall-signal channel is excluded from geometric transforms since its $12\times8$ temporal encoding is position-dependent.

All model variants, including the baseline, are trained and evaluated on identical data splits under an identical training pipeline, so that every comparison in this paper isolates the loss function alone. The primary configuration is a shuffled 80/20 train/validation split (240/60~samples, seed~42). Two forms of robustness testing are added: (i)~\hl{split-repeat} %
 runs, in which the key variants are retrained on two additional shuffled splits (seeds~7 and 123) and results are reported as mean~$\pm$~standard deviation across the three splits; and (ii)~\hl{leave-one-storm-out} (LOSO) runs, in which a variant is trained on one storm event and evaluated on all 150~samples of the held-out storm (10\% of the training storm is retained for early stopping), testing generalization to an unseen event.

\subsection{Training Configuration}

The training hyperparameters are summarized in Table~\ref{tab:config}. The simulations were performed using TensorFlow (v2.21.0) with the Keras~3 API (v3.15.0) under Python (v3.12.11) on a single NVIDIA RTX~4000~Ada GPU (NVIDIA Corporation, Santa Clara, CA, USA). Each variant trains in under ten minutes, consistent with the real-time deployability motivating the surrogate approach.

\begin{table}[H]
    \caption{Training hyperparameters and reproducibility details. Actual epochs trained before early stopping are reported per variant in Section~\ref{sec:training_behavior}.}
    \label{tab:config}
\begin{tabular}{ll}
        \toprule
        \textbf{Parameter} & \textbf{Value}\\
        \midrule
        Optimizer & Adam\\
        Initial learning rate & $5 \times 10^{-5}$\\
        Batch size & 4\\
        Max epochs & 50 (early stopping, patience\,=\,15)\\
        LR schedule & ReduceLROnPlateau (factor\,=\,0.5, patience\,=\,5)\\
        Backbone & ResNet50 (ImageNet pretrained, early layers frozen)\\
        Physics warm-up & 15 epochs (linear ramp)\\
        Gravity target weight & 1.0\\
        Continuity target weight & 0.1\\
        Continuity $\epsilon_{\text{base}}$, $\alpha$ & 0.01, 0.5 (normalized depth units)\\
        Parameters & 40.1\,M total (38.7\,M trainable)\\
        Training time per epoch & 5--6\,s (RTX 4000 Ada, batch size 4)\\
        \bottomrule
\end{tabular}
\end{table}

\subsection{Model Variants}

Five model variants are evaluated to isolate the contribution of each loss component, all sharing the identical architecture of Section~\ref{sec:method}:

\begin{itemize}[leftmargin=*,labelsep=5.8mm]
    \item \hl{Baseline (V0):} %
 The architecture of Wang et al.~\cite{wang2026hybrid}, trained with standard MSE loss. No physics constraints.
    \item \hl{V1 (Physics-Informed):} $10\times$ weighted MSE + gravity loss + continuity loss with 15-epoch warm-up.
    \item \hl{V2 (High Wet Weight):} $100\times$ weighted MSE + gravity + continuity. An ablation testing aggressive class-imbalance correction.
    \item \hl{V3 (False Alarm + Sparsity):} $10\times$ weighted MSE + $20\times$ uniform false alarm + $50\times$ sparsity constraint + gravity + continuity.
    \item \hl{V4 (TWI False Alarm):} Identical to V3 except the uniform false-alarm penalty is replaced by the TWI-modulated penalty of Equation~\eqref{eq:twi_fa} at the same base weight ($20\times$), so that the only difference between V3 and V4 is the topographic modulation.
\end{itemize}

In addition, a \hl{generic-physics comparator} is trained ($10\times$ weighted MSE + temporal-acceleration and spatial Laplacian smoothness penalties), representing physics-agnostic regularization of the kind used in generic PINN-style formulations. This comparator tests whether the benefits of the proposed losses stem from their hydrology-specific structure rather than from regularization per se.

\subsection{Quantitative Evaluation}

Table~\ref{tab:results} presents per-timestep RMSE and MAE for each model variant on the primary split. Unlike an earlier version of this study, the baseline is retrained on the identical split and pipeline rather than compared against the aggregate statistics reported by \mbox{Wang et al.~\cite{wang2026hybrid}}, eliminating split mismatch as a confounder. Notably, the retrained baseline is strong: V0 attains the best aggregate error of all variants (MAE 0.009\,m, RMSE 0.032\,m), which we attribute to the shared training improvements (ImageNet initialization, augmentation, unbounded ReLU output head) benefiting the unconstrained objective most. Among the constrained variants, V4 achieves the lowest MAE (0.013\,m), improving on both V1 (0.016\,m) and V3 (0.017\,m on this split), while V1 attains the lowest RMSE (0.035\,m). As shown in Section~\ref{sec:street_eval}, aggregate error tells only half the story: the variants differ far more in \hl{where} they place water than in how much error they accumulate.

The gravity loss of every constrained variant converges to $\sim$10$^{-6}$ or below (V1: $7.7\times10^{-7}$ training, $1.1\times10^{-6}$ validation), effectively zero, confirming that the models successfully learn not to push water uphill. Continuity losses stabilize at $\sim$2--7$\times10^{-5}$.

\begin{figure}[H]
        \includegraphics[width=.99\textwidth]{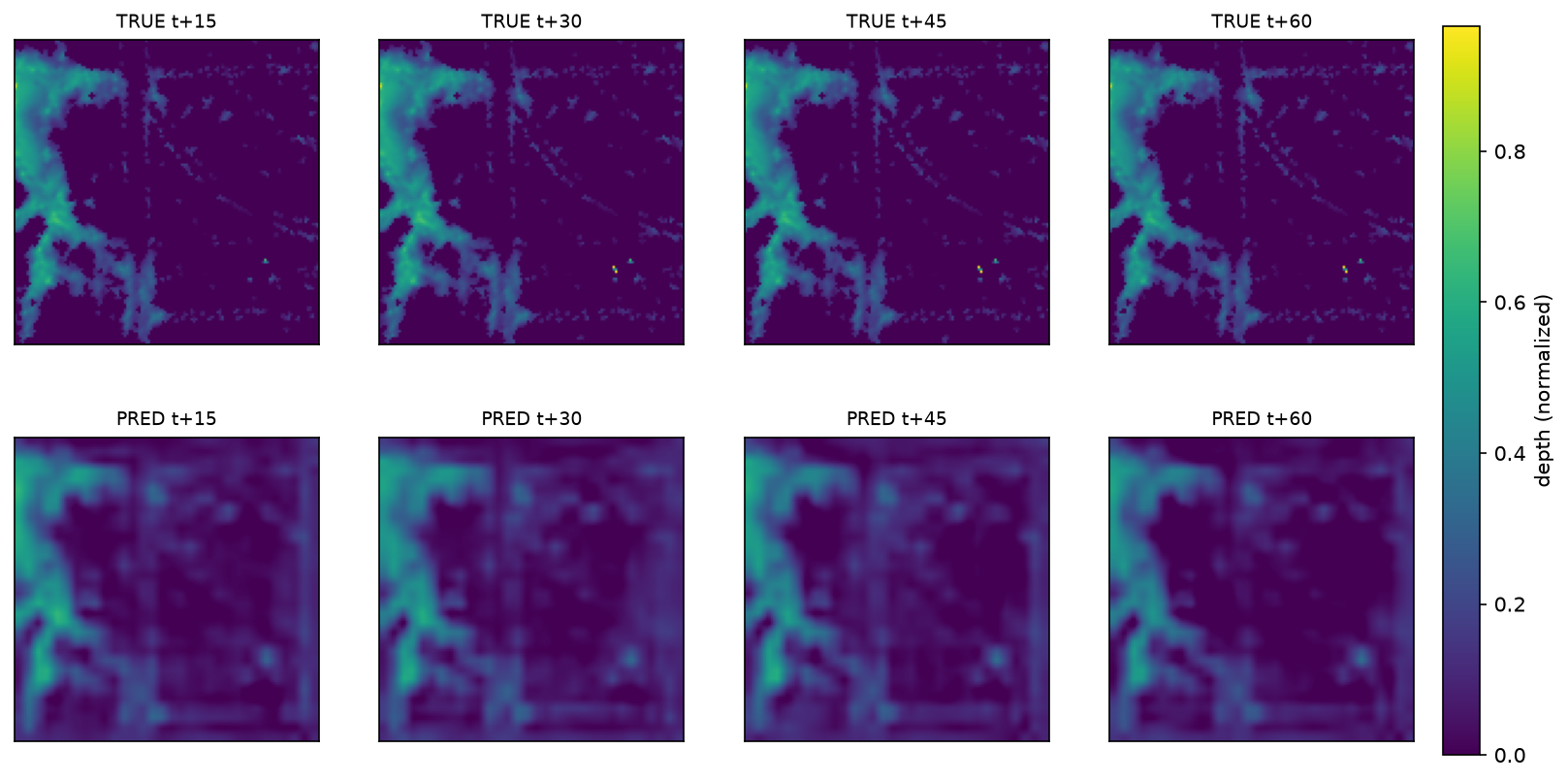}
    \caption{V2 ($100\times$ wet pixel weight) failure mode. Ground truth (\textbf{top}) vs.\ predictions (\textbf{bottom}) for the 30~September~2022 event. The model spreads low-magnitude water across the full grid (uniform bluish wash) rather than localizing predictions to flood corridors, with a single bright spot emerging at the location of the strongest ground-truth peak. This ``hedging'' behavior, predicting some water everywhere to avoid the asymmetric cost of missing a wet pixel, motivated abandoning V2 and introducing explicit false-alarm and sparsity penalties in V3. Ground-truth and prediction panels share the same color scale.}
    \label{fig:v2_failure}
\end{figure}

\begin{figure}[H]
        \includegraphics[width=\textwidth]{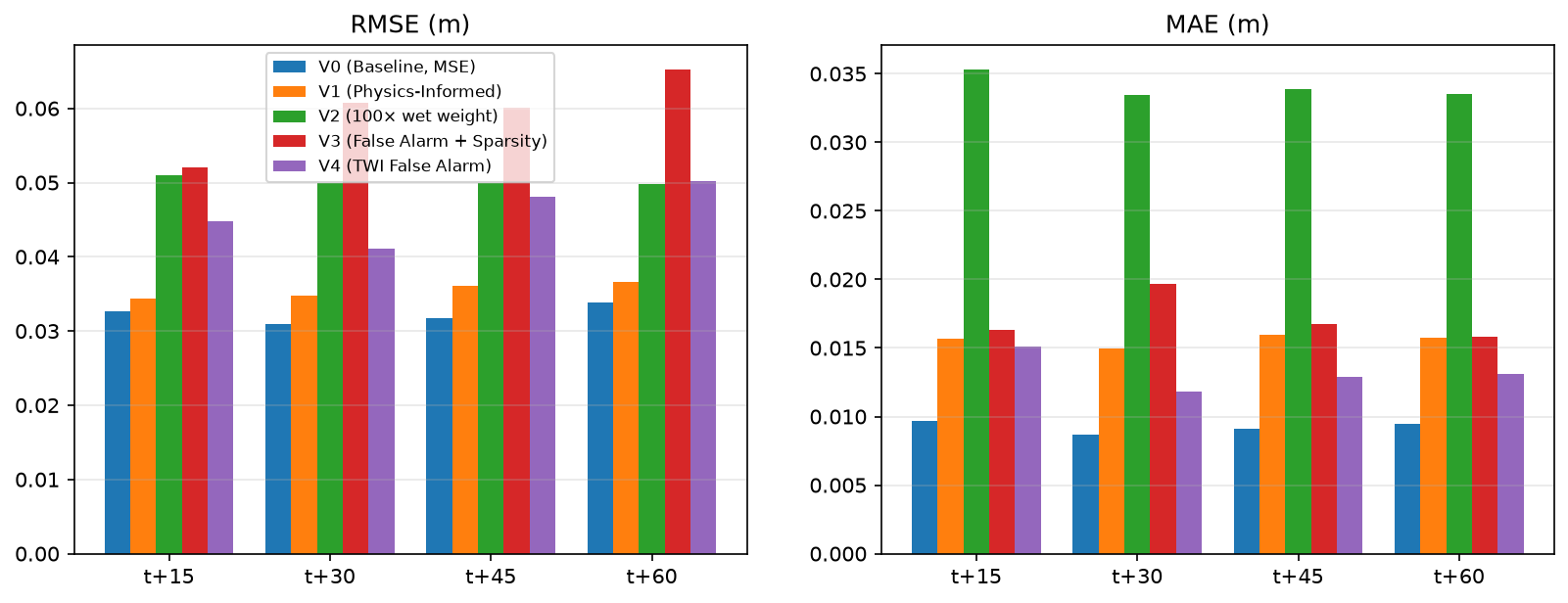}
    \caption{\hl{RMSE} %
 (\textbf{left}) and MAE (\textbf{right}) per forecast timestep for all five model variants on the identical validation split. V2 ($100\times$ wet weight) is roughly $2$--$3\times$ worse than the other constrained variants on both metrics across every horizon, illustrating the failure of naive class-imbalance correction. \mbox{V0 achieves} the best aggregate accuracy; among constrained variants, V4 attains the lowest MAE and V1 the lowest RMSE.}
    \label{fig:bar_chart}
\end{figure}

V2 ($100\times$ wet weight) reproduces the diffuse-wash failure mode: the model spreads thin water predictions across the grid rather than localizing them (Figure~\ref{fig:v2_failure}), predicting a 69\% wet pixel fraction against the 9\% present in the validation ground truth. Quantitatively V2 is the worst variant by a wide margin (MAE 0.034\,m, roughly $2\times$ worse than V1 and $2.6\times$ worse than V4), a gap that persists across every forecast horizon on both metrics (Figure~\ref{fig:bar_chart}). With extreme wet pixel weighting, the model found it optimal to hedge by predicting low water everywhere rather than risk missing any wet pixel. Notably, V2's physics losses remained stable throughout training (gravity $1.2\times10^{-6}$, continuity $7.3\times10^{-5}$ at convergence; Section~\ref{sec:training_behavior}), demonstrating that the physics terms are robust to aggressive data-loss reweighting: the failure is driven entirely by the weighted MSE term.

The generic-physics comparator achieves competitive aggregate MAE (0.016\,m) but, as shown below, collapses street-level skill almost entirely (street recall 0.01), confirming that generic smoothness regularization is no substitute for hydrology-specific structure.

  \vspace{-3pt}   

  \vspace{-9pt}   
  
\begin{table}[H]
    \caption{\hl{Per-timestep} %
 error metrics across all model versions, measured on the identical 60-sample validation split (seed 42) under the identical training pipeline. All values in meters (normalized depth units); best values in bold; best among constrained variants underlined.}
    \label{tab:results}
        
\begin{adjustwidth}{-\extralength}{0cm}

\footnotesize\setlength{\tabcolsep}{3pt}%
\begin{tabular}{p{5.24cm}cccccccccc}
         \toprule
        & \multicolumn{4}{c}{\textbf{RMSE (m)}}
        & \multicolumn{4}{c}{\textbf{MAE (m)}}
        & \multicolumn{2}{c}{\textbf{Average}}\\
        \cmidrule(lr){2-11}
        \textbf{Model} & \textbf{\boldmath{$t\!+\!15$}} & \textbf{\boldmath{$t\!+\!30$}} & \textbf{\boldmath{$t\!+\!45$}} & \textbf{\boldmath{$t\!+\!60$}}
                       & \textbf{\boldmath{$t\!+\!15$}} & \textbf{\boldmath{$t\!+\!30$}} & \textbf{\boldmath{$t\!+\!45$}} & \textbf{\boldmath{$t\!+\!60$}}
                       & \textbf{RMSE} & \textbf{MAE}\\
        \midrule
        V0 (Baseline, MSE)                & \textbf{0.033} & \textbf{0.031} & \textbf{0.032} & \textbf{0.034}
                                          & \textbf{0.010} & \textbf{0.009} & \textbf{0.009} & \textbf{0.009}
                                          & \textbf{0.032} & \textbf{0.009}\\
        V1 (Physics-Informed)             & \underline{0.034} & \underline{0.035} & \underline{0.036} & \underline{0.037}
                                          & 0.016 & 0.015 & 0.016 & 0.016
                                          & \underline{0.035} & 0.016\\
        V2 ($100\times$ Wet Weight)       & 0.051 & 0.050 & 0.050 & 0.050
                                          & 0.035 & 0.033 & 0.034 & 0.034
                                          & 0.050 & 0.034\\
        V3 (False Alarm + Sparsity)       & 0.052 & 0.061 & 0.060 & 0.065
                                          & 0.016 & 0.020 & 0.017 & 0.016
                                          & 0.060 & 0.017\\
        V4 (TWI False Alarm)              & 0.045 & 0.041 & 0.048 & 0.050
                                          & \underline{0.015} & \underline{0.012} & \underline{0.013} & \underline{0.013}
                                          & 0.046 & \underline{0.013}\\
        \midrule
        Generic-Physics Comparator        & 0.063 & 0.064 & 0.065 & 0.065
                                          & 0.016 & 0.015 & 0.016 & 0.016
                                          & 0.064 & 0.016\\
        \bottomrule
\end{tabular}
  \end{adjustwidth}
    \end{table}

\subsection{Street-Level Evaluation}
\label{sec:street_eval}

Aggregate pixel-averaged metrics cannot reveal whether a model captures the thin, safety-critical street corridors that motivate this work. A road-proximal evaluation protocol is therefore introduced. A \hl{street mask} is derived from the DEM as the set of pixels with TWI$_{\text{norm}} \geq 0.7$ (Section~\ref{sec3.3.4}), identifying the valley and street-channel pixels where water plausibly accumulates; no road-network data is required. Within this mask, a pixel is counted as flooded when its depth exceeds 0.01 normalized units, and recall (fraction of truly flooded street pixels predicted flooded), precision (fraction of predicted flooded street pixels that are truly flooded), and their harmonic mean F1 are computed over all validation samples and forecast horizons. The results are insensitive to the mask threshold: varying it between 0.6 and 0.8 changes no value by more than 0.001.

Table~\ref{tab:street} quantifies what Section~\ref{sec:qualitative} shows visually. V1 captures 85\% of truly flooded street pixels, against 54\% for the baseline on this split (0.44~$\pm$~0.10 across splits; Section~\ref{sec:robustness}), at the cost of over-prediction (precision 0.26). V3's uniform false-alarm penalty suppresses street recall to 0.17, the collapse that aggregate MAE conceals. V4 recovers $2.2\times$ higher recall than V3 while also improving precision, yielding the best F1 among all constrained variants, and does so at the lowest constrained-variant MAE (Table~\ref{tab:results}). The generic-physics comparator, despite competitive aggregate MAE, is unusable for street-level prediction (recall 0.01): smoothness regularization actively removes exactly the thin structures the application requires.

\begin{table}[H]
    \caption{\hl{Street-level} %
 metrics on the primary validation split (seed 42). Recall is the safety-critical quantity for traffic routing: a missed flooded street pixel corresponds to routing traffic into water. Best values in bold; best among constrained variants underlined.}
    \label{tab:street}
   \begin{tabularx}{\textwidth}{lCCC}
        \toprule
        \textbf{Model} & \textbf{Street Recall} & \textbf{Street Precision} & \textbf{Street F1}\\
        \midrule
        V0 (Baseline, MSE)          & 0.545 & \textbf{0.580} & \textbf{0.561}\\
        V1 (Physics-Informed)       & \underline{\textbf{0.849}} & 0.261 & 0.400\\
        V2 ($100\times$ Wet Weight) & 0.979 $^{\dagger}$ & 0.129 & 0.228\\
        V3 (False Alarm + Sparsity) & 0.174 & 0.200 & 0.186\\
        V4 (TWI False Alarm)        & 0.390 & \underline{0.469} & \underline{0.426}\\
        \midrule
        Generic-Physics Comparator  & 0.013 & 0.297 & 0.026\\
        \bottomrule
        \end{tabularx}
        
       \noindent{\footnotesize{Note: $^{\dagger}$ V2's near-perfect recall is an artifact of predicting water almost everywhere (69\% wet fraction).}}

\end{table}

\subsection{Robustness: Split Repeats and Cross-Storm Generalization}
\label{sec:robustness}

Table~\ref{tab:robustness} reports the baseline and key variants retrained across three shuffled splits and in both leave-one-storm-out directions. Three observations follow. First, the variant ordering is stable across every split and both LOSO directions: V1 always attains the highest street recall, V3 always the lowest, and V4 always lies between them on recall while achieving the best MAE and F1 among constrained variants; the comparison is therefore a property of the loss formulations, not an artifact of a particular split. Second, the baseline's street recall is itself split-sensitive, ranging from 0.31 to 0.54 across seeds (0.44~$\pm$~0.10), whereas its aggregate error is highly stable; aggregate metrics mask the variability of exactly the application-critical skill. Third, the physics advantage \emph{widens} under cross-storm evaluation: on the unseen 2022 storm the baseline's street recall falls to 0.28 while V1 retains 0.61, more than double, and V1 also achieves the better RMSE in that direction. Physics constraints thus function as an inductive bias that transfers to unseen events, where purely statistical street-level skill degrades most.

\begin{table}[H]
    \caption{Robustness of the baseline and key variants. Top: Mean $\pm$ standard deviation over three shuffled 80/20 splits (seeds 42, 7, 123). Bottom: Leave-one-storm-out (LOSO), trained on one storm and evaluated on all 150 samples of the held-out storm. Best per column in bold.}
    \label{tab:robustness}

  \begin{adjustwidth}{-\extralength}{0cm}
\footnotesize\setlength{\tabcolsep}{3pt}%
\begin{tabular}{p{5.5cm}lcccc}
        \toprule
        \textbf{Evaluation} & \textbf{Model} & \textbf{MAE (m)} & \textbf{RMSE (m)} & \textbf{Street Recall} & \textbf{Street F1}\\
        \midrule
        \multirow{4}{*}{3 random splits (mean $\pm$ std)}
        & V0 & \textbf{0.0086 $\pm$ 0.0008} & \textbf{0.0323 $\pm$ 0.0010} & 0.442 $\pm$ 0.100 & \textbf{0.494 $\pm$ 0.061}\\
        & V1 & 0.0165 $\pm$ 0.0016 & 0.0377 $\pm$ 0.0019 & \textbf{0.770 $\pm$ 0.093} & 0.347 $\pm$ 0.039\\
        & V3 & 0.0138 $\pm$ 0.0025 & 0.0489 $\pm$ 0.0083 & 0.246 $\pm$ 0.140 & 0.260 $\pm$ 0.150\\
        & V4 & 0.0125 $\pm$ 0.0013 & 0.0426 $\pm$ 0.0052 & 0.393 $\pm$ 0.121 & 0.385 $\pm$ 0.099\\
        \midrule
        \multirow{4}{*}{LOSO: train 2017 $\rightarrow$ test 2022}
        & V0 & \textbf{0.0172} & 0.0607 & 0.281 & 0.372\\
        & V1 & 0.0201 & \textbf{0.0545} & \textbf{0.613} & \textbf{0.419}\\
        & V3 & 0.0255 & 0.0729 & 0.259 & 0.226\\
        & V4 & 0.0192 & 0.0593 & 0.402 & 0.394\\
        \midrule
        \multirow{4}{*}{LOSO: train 2022 $\rightarrow$ test 2017}
        & V0 & \textbf{0.0104} & \textbf{0.0381} & 0.398 & \textbf{0.400}\\
        & V1 & 0.0163 & 0.0421 & \textbf{0.606} & 0.307\\
        & V3 & 0.0146 & 0.0474 & 0.177 & 0.143\\
        & V4 & 0.0131 & 0.0422 & 0.391 & 0.325\\
        \bottomrule
\end{tabular}
	\end{adjustwidth}
	
\end{table}

\subsection{Qualitative Evaluation}
\label{sec:qualitative}

\begin{figure}[H]
        \includegraphics[width=.99\textwidth]{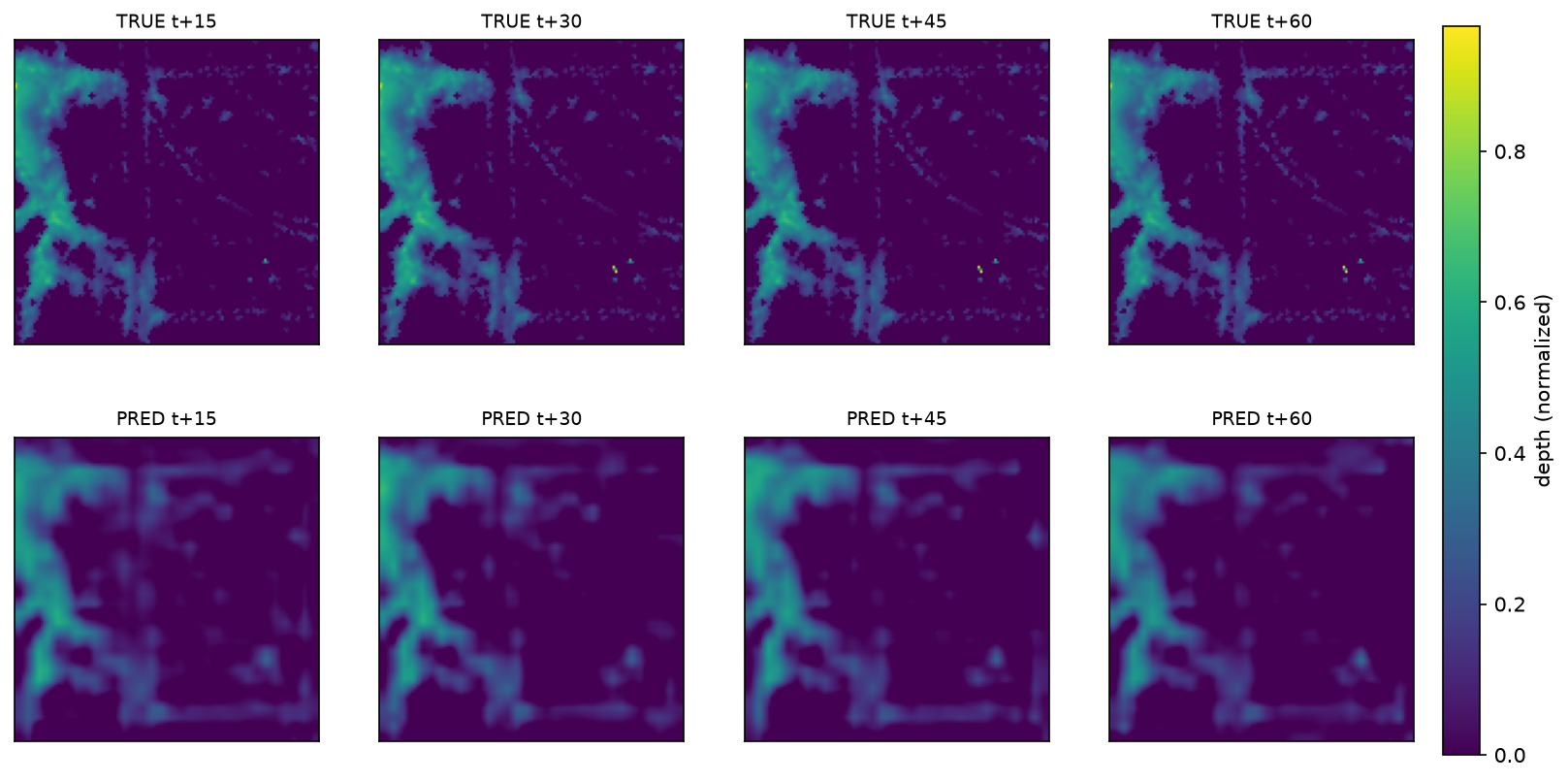}
    \caption{V1 (Physics-Informed): Ground truth (\textbf{top}) vs.\ predictions (\textbf{bottom}) for the 30~September~2022 event (gauge reading: 65.75\,ft). Columns correspond to the four 15 min forecast horizons $t\!+\!15$, $t\!+\!30$, $t\!+\!45$, and $t\!+\!60$. Note the thin linear flood corridors along street channels. Ground-truth and prediction panels share the same color scale.}
    \label{fig:v1_pred}
\end{figure}

\begin{figure}[H]
        \includegraphics[width=.99\textwidth]{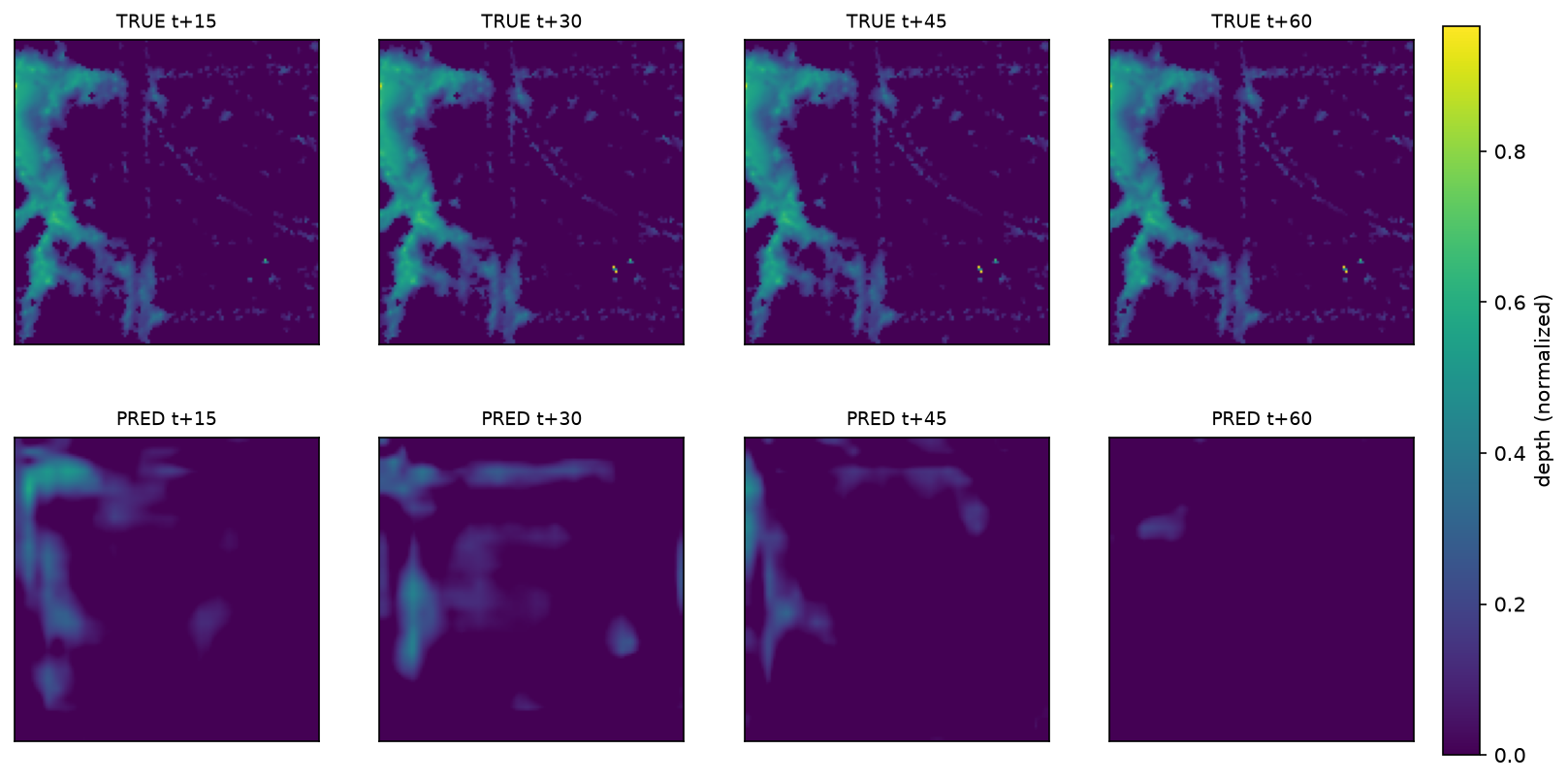}
    \caption{V3 (False Alarm + Sparsity): Ground truth (\textbf{top}) vs.\ predictions (\textbf{bottom}) for the same event and timesteps. Street-channel flooding is suppressed by the uniform false-alarm penalty, and the suppression intensifies with forecast horizon. Ground-truth and prediction panels share the same color scale.}
    \label{fig:v3_pred}
\end{figure}

\begin{figure}[H]
        \includegraphics[width=.99\textwidth]{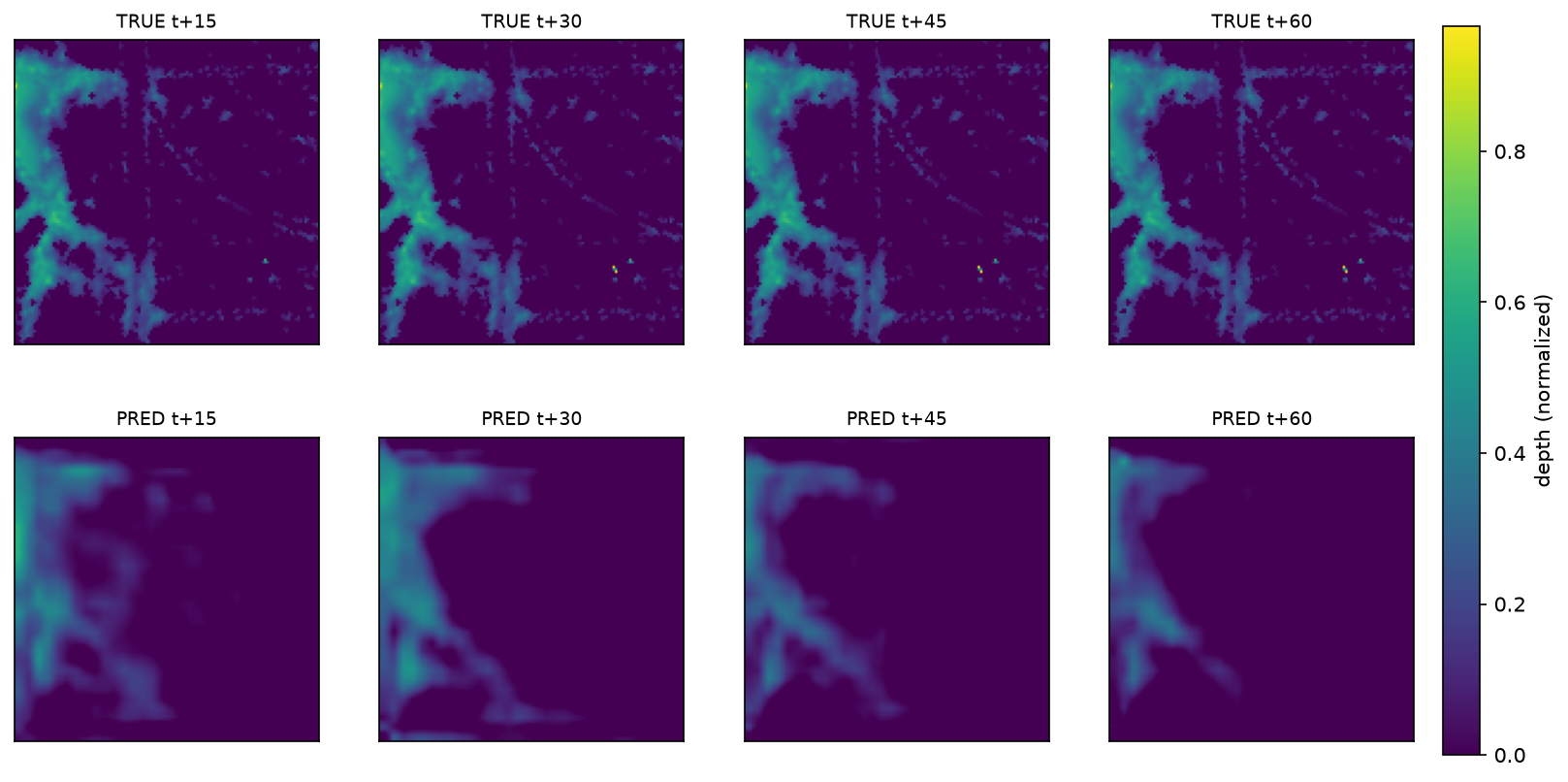}
    \caption{\hl{V4 (TWI False Alarm):} %
 Ground truth (\textbf{top}) vs.\ predictions (\textbf{bottom}) for the same event and timesteps. The TWI-modulated penalty preserves the main flood structure at every horizon with a background as clean as V3's, avoiding V3's progressive collapse. Ground-truth and prediction panels share the same color scale.}
    \label{fig:v4_pred}
\end{figure}

\begin{figure}[H]

\begin{adjustwidth}{-\extralength}{0cm}
\centering 
        \includegraphics[width=\textwidth]{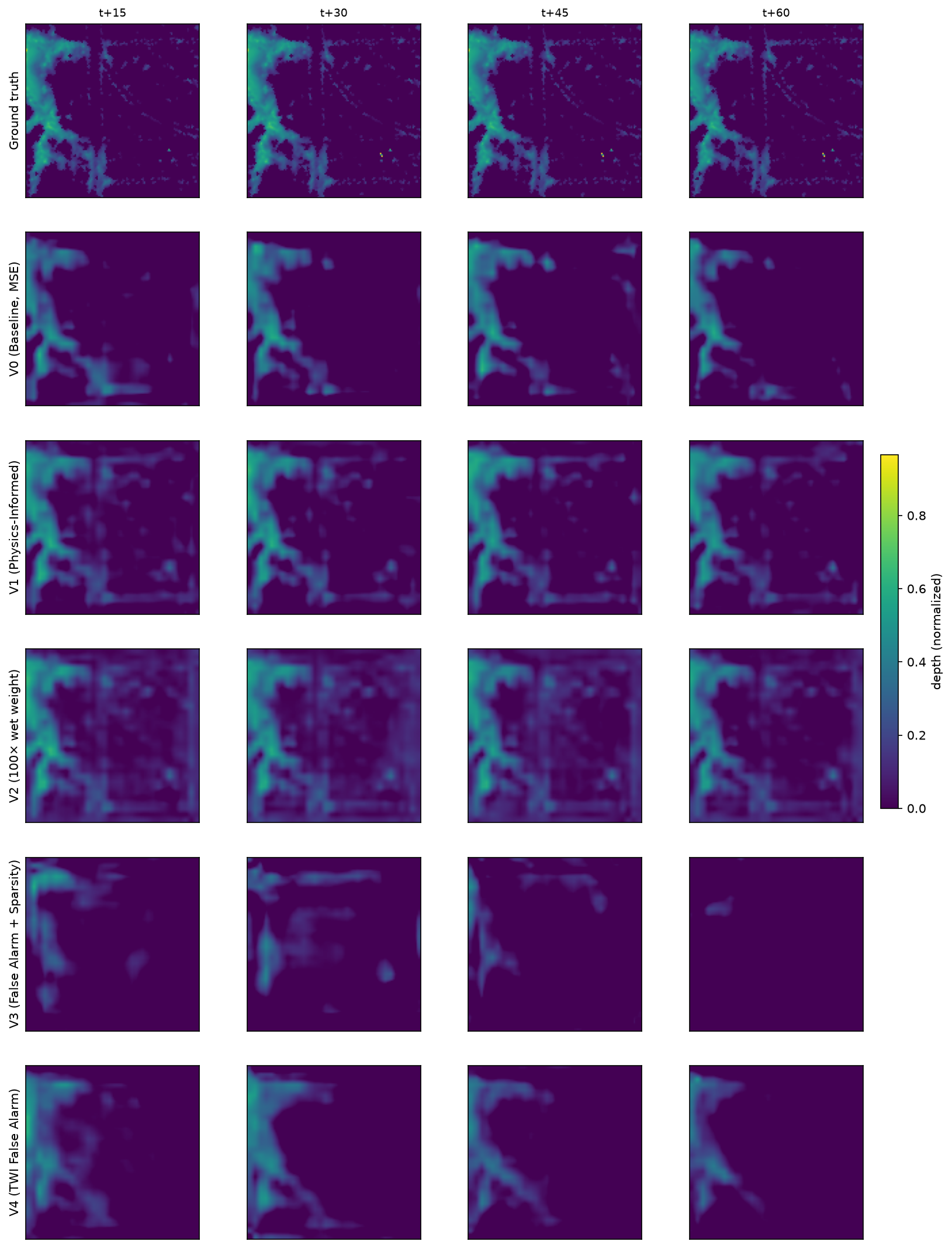}
\end{adjustwidth}
    \caption{Side-by-side ablation across all five model variants for the 30~September~2022 event. Rows from top: Ground truth, V0 baseline (Wang et al.~\cite{wang2026hybrid} architecture + MSE), V1 (physics-informed), \mbox{V2 ($100\times$ wet weight),} V3 (false alarm + sparsity), V4 (TWI false alarm). Columns correspond to the four 15 min forecast horizons. V0 captures the main flood body but omits most thin street corridors; V1 recovers the corridors with some over-spreading; V2 exhibits the diffuse-wash failure mode; V3 suppresses flooding progressively with horizon; V4 preserves the main flood structure with a clean background. All panels share the same color scale.}
    \label{fig:all_versions}
\end{figure}

Qualitative inspection reveals differences in prediction behavior that the aggregate metrics conceal. Figures~\ref{fig:v1_pred}--\ref{fig:v4_pred} show V1, V3, and V4 predictions for the 30~September~2022 storm event alongside the ground truth, and \hl{Figure}~\ref{fig:all_versions} %
 compares all five variants.

\begin{figure}[H]

\begin{adjustwidth}{-\extralength}{0cm}
\centering 
        \includegraphics[width=\textwidth]{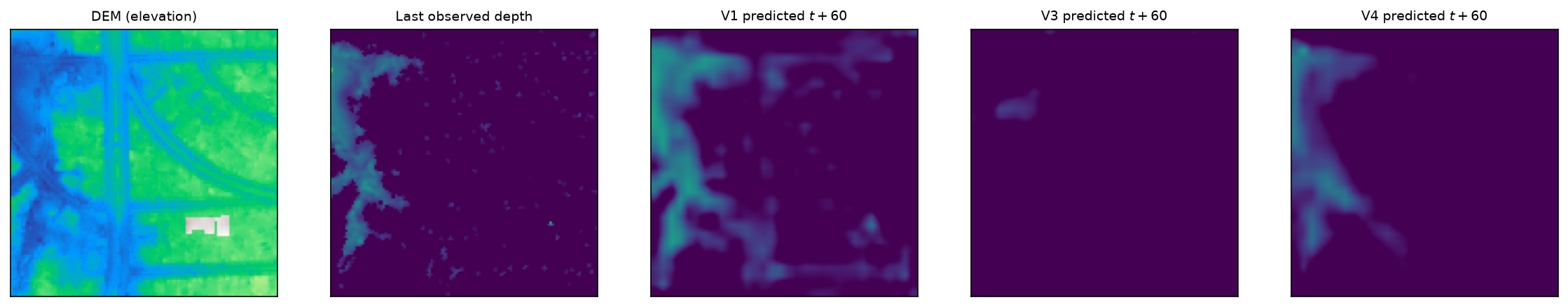}
\end{adjustwidth}
    \caption{Input context and model comparison. From left: DEM (elevation), last observed depth, V1 predicted $t\!+\!60$, V3 predicted $t\!+\!60$, V4 predicted $t\!+\!60$. V1's flood corridors align with the low-elevation road network visible in the DEM; V3 suppresses these channels; V4 preserves the main flood structure with a clean background.}
    \label{fig:context}
\end{figure}

The ground truth contains a main flood body along the western drainage corridor and a network of thin, dotted street corridors across the eastern half of the tile. V0 reproduces the main flood body cleanly but largely omits the thin eastern corridors, consistent with its street recall of 0.54. V1 recovers substantially more of these corridors, narrow, linear features that align with the low-elevation road network visible in the DEM (\hl{Figure}~\ref{fig:context}), at the cost of some over-spreading around flood boundaries (precision 0.26). The gravity loss, by penalizing water at topographically high pixels, channels predictions into these low-elevation corridors. This topographic awareness is the defining qualitative advantage of the physics-informed approach.

V3 suppresses street-channel predictions severely, and the suppression compounds with forecast horizon: its $t\!+\!60$ prediction retains almost no flooded area (Figure~\ref{fig:v3_pred}), consistent with its street recall of 0.17. Its uniform false-alarm penalty treats all dry pixels equally: it cannot distinguish a narrow road channel (where flooding is physically plausible and safety-critical) from a localized highpoint or ridgeline (where it is not). Streets are topographically narrow and linear; the false-alarm penalty classifies them as noise and suppresses both.

V4 (Figure~\ref{fig:v4_pred}) preserves the main flood structure across all four horizons with a background as clean as V3's, visibly avoiding both V2's wash and V3's progressive collapse. Its recovery of the thinnest eastern corridors is partial, reflected in its street recall of 0.39 versus V1's 0.85, but its predictions remain usable at $t\!+\!60$ where V3's have vanished.

  \vspace{-3pt}   
  \vspace{-9pt}    

\subsection{Training Behavior and Overfitting Control}
\label{sec:training_behavior}

\begin{figure}[H]

\begin{adjustwidth}{-\extralength}{0cm}
\centering 
        \includegraphics[width=\textwidth]{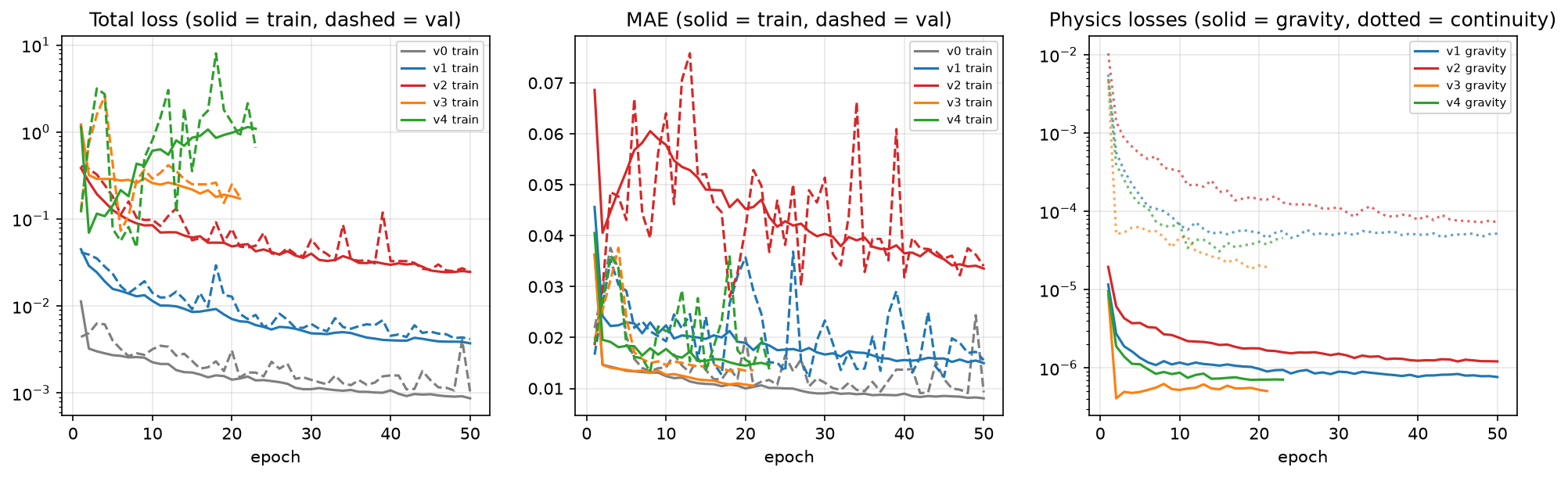}
\end{adjustwidth}
    \caption{Training dynamics on the primary split. \textbf{Left}: Total loss (solid = training, dashed = validation; log scale). \textbf{Center}: MAE. \textbf{Right}: Physics loss components (solid = gravity, dotted = continuity; log scale). Validation curves track training curves without divergence, and physics losses converge stably for all constrained variants, including V2 under $100\times$ wet weighting.}
    \label{fig:loss_curves}
\end{figure}

Figure~\ref{fig:loss_curves} presents training and validation trajectories for all variants, addressing two questions: whether a 40.1\,M-parameter ResNet50-based model overfits 240 training samples, and how the physics losses behave during optimization.

Validation loss tracks training loss without divergence for all variants; the mitigations are ImageNet initialization with frozen early layers (only 38.7\,M of 40.1\,M parameters are trainable, and the frozen layers carry generic features), flip/rotation augmentation, and early stopping. V0, V1, and V2 train for the full 50 epochs; V3 and V4 stop early at epochs 21 and 23 respectively (best validation loss at epochs 6 and 8), reflecting the stronger regularization of their structural penalties.

The physics loss trajectories confirm stable optimization. Gravity losses for all constrained variants decay monotonically after the 15-epoch warm-up and converge to $\sim$10$^{-6}$ or below without oscillation. Notably, V2's trajectories remain as stable as V1's despite the $100\times$ wet weighting, converging to gravity $1.2\times10^{-6}$ and continuity $7.3\times10^{-5}$: the physics terms are robust to aggressive data-loss reweighting, localizing V2's diffuse-wash failure entirely in the weighted MSE term.

\subsection{The Quantitative--Qualitative Trade-Off}

Table~\ref{tab:summary} summarizes the trade-off. The baseline attains the best aggregate error; V1 provides the strongest street-channel capture; the uniform false-alarm penalty (V3) buys clean backgrounds at the cost of collapsing the safety-critical corridors; and the TWI-modulated penalty (V4) improves on V3 in every measured dimension. This divergence arises because aggregate pixel-level error is an imperfect proxy for the downstream task:

\begin{itemize}[leftmargin=*,labelsep=5.8mm]
    \item A correct prediction of 0.15\,m flooding on a specific road segment has high application value, even if it affects only a thin corridor of pixels.
    \item A missed prediction of that same corridor contributes marginally to aggregate MAE (few pixels) but is critical to the application (vehicles enter a flooded road).
\end{itemize}

  \vspace{-9pt}   
  \vspace{-9pt}    

  \vspace{-6pt}    

\begin{table}[H]
    \caption{Summary comparison across all five model versions on the identical validation split (seed 42). Percentages are relative to the retrained V0 baseline. Gravity loss is the converged training value; the baseline does not include the term. Best values in bold; best among constrained variants underlined.}
    \label{tab:summary}
    \begin{adjustwidth}{-\extralength}{0cm}
\footnotesize\setlength{\tabcolsep}{3pt}%
\begin{tabular}{lccccc}
        \toprule
        \textbf{Metric} & \textbf{V0 Baseline} & \textbf{V1 Physics} & \textbf{V2 Diffuse} & \textbf{V3 Uniform FA} & \textbf{V4 TWI FA}\\
        \midrule
        MAE (m)                 & \textbf{0.009} & 0.016 ($+$69\%) & 0.034 ($+$269\%) & 0.017 ($+$86\%) & \underline{0.013} ($+$44\%)\\
        RMSE (m)                & \textbf{0.032} & \underline{0.035} ($+$10\%) & 0.050 ($+$55\%) & 0.060 ($+$84\%) & 0.046 ($+$42\%)\\
        Wet pixel fraction (truth: 9.3\%) & \textbf{8.5\%} & 29.6\% & 69.0\% (wash) & 7.9\% & \underline{7.6\%}\\
        Street recall           & 0.545 & \underline{\textbf{0.849}} & 0.979 (hedged) & 0.174 & 0.390\\
        Street F1               & \textbf{0.561} & 0.400 & 0.228 & 0.186 & \underline{0.426}\\
        Gravity loss (converged) & --- & $7.7\times10^{-7}$ & $1.2\times10^{-6}$ & $5.1\times10^{-7}$ & $7.1\times10^{-7}$\\
        \bottomrule
        
\end{tabular}
	\end{adjustwidth}
\end{table}

\section{Discussion}
\label{sec:discussion}

\subsection{The Gravity Loss as an Implicit Road Detector}

The gravity constraint is the single most impactful component for road-level flood recall. By penalizing water at topographically high elevations, it naturally channels predictions into low-lying road networks, without any explicit road-geometry encoding: V1 recalls 77--85\% of truly flooded street pixels against the baseline's 44\% $\pm$ 10\%, and retains a $>$2$\times$ recall advantage on a held-out storm where the baseline's street skill degrades most (Section~\ref{sec:robustness}). Streets, being the lowest points in urban terrain grids, become the preferred locations for model predictions. This emergent behavior is particularly valuable because road-network data is not required as an input. It is noteworthy that the retrained baseline itself is a strong model, achieving the best aggregate error and, through its balanced recall/precision, the best street F1; the contribution of the physics constraints is specifically the large recall gain, which is the safety-critical direction for routing, where a missed flooded street is more costly than a false alarm that can be cleared by slower verification.

\subsection{Aggregate Metrics Can Mislead}

The V1/V3/V4 comparison demonstrates that conventional evaluation metrics, pixel-averaged MAE and RMSE, can favor models that perform worse on application-critical features: on averaged splits V3 attains 16\% lower MAE than V1 while capturing less than a third as many flooded street pixels. For traffic routing, per-road-segment accuracy matters more than spatially averaged error. The road-proximal recall/precision/F1 protocol of Section~\ref{sec:street_eval} operationalizes this observation and revealed a collapse (V3 recall 0.17) that was invisible in Table~\ref{tab:results}; we argue such task-specific metrics should accompany aggregate error in flood-surrogate evaluation generally.

\subsection{Naive Class-Imbalance Correction Is Counterproductive}

The V2 failure mode ($100\times$ wet weight producing diffuse wash) illustrates that aggressive reweighting without structural penalties can be counterproductive. The model discovered that minimizing expected loss under extreme wet pixel weighting is achieved by hedging, predicting low water everywhere rather than risking high per-pixel error at any wet location. That V2's physics losses remained stable throughout (Section~\ref{sec:training_behavior}) shows the pathology is not an interaction effect: reweighting alone is the culprit, and structural penalties, not stronger weights, are the appropriate correction.

\subsection{TWI Modulation Reconciles the Trade-Off}

The TWI-modulated false-alarm penalty (Equation~\eqref{eq:twi_fa}) empirically reconciles the quantitative--qualitative trade-off. Because V4 differs from V3 only in the topographic modulation, the comparison isolates its effect: V4 improves on V3 in MAE, RMSE, street recall ($+60\%$ relative), street precision, and street F1, simultaneously, across all three random splits and both leave-one-storm-out directions. By reducing false-alarm penalties in topographic valleys while retaining them on elevated surfaces, the modulation permits street-channel predictions that the uniform penalty erases, while preserving its noise suppression on non-physical locations. The reconciliation is partial rather than total: V4's street recall (0.39) remains well below V1's (0.85), so applications that prioritize recall above all else should still prefer the pure physics formulation, while applications requiring balanced extent estimates are best served by V4. The TWI is computed directly from the DEM, which is already available as input channel~0, so the improvement is architecture- and data-free.

\section{Conclusions}
\label{sec:conclusion}

In this work, a physics-informed loss function framework for CNN-LSTM urban flood forecasting has been presented that embeds gravitational flow, mass conservation, and topography-aware spatial constraints into the training objective. Experiments on the Norfolk, Virginia, flood dataset, with all variants trained on identical splits and validated across repeated splits and leave-one-storm-out tests, reveal that physics constraints, particularly the gravity loss, substantially raise street-channel recall (0.77 $\pm$ 0.09 versus 0.44 $\pm$ 0.10 for the retrained baseline, widening to a $>$2$\times$ advantage on a held-out storm), the capability most relevant to traffic-routing applications, at a modest cost in aggregate error.

The central finding of this work is the tension between aggregate pixel-level accuracy and application-specific physical plausibility, and its resolution through terrain-aware loss modulation. A uniform false-alarm penalty (V3) achieves clean predictions whose street-level recall collapses to 0.17; replacing it with the TWI-modulated penalty (V4), a change in a single multiplicative field derived from the DEM, improves every metric measured, yielding the best F1 and lowest MAE among constrained variants across all splits and both held-out storms. A model optimized for pixel-averaged error and a model optimized for road-segment recall remain different products, but physics-informed loss functions, and topographically modulated penalties in particular, provide a mechanism to steer this trade-off without modifying the model architecture.

In future work, further evaluation will be carried out with different configurations:

\begin{enumerate}[leftmargin=*,labelsep=4.9mm]
    \item \hl{Closing the recall gap:} %
 V4 recovers street corridors only partially (recall 0.39 versus V1's 0.85); stronger valley relief (lower TWI multiplier floors), or combining TWI modulation with recall-oriented terms, may close the remaining gap.
    \item \hl{Two-head architecture:} Decomposing the prediction into a binary classification head (flood extent) and a regression head (flood depth), with final output as their product, to enable hard spatial decisions before magnitude estimation.
    \item \hl{Traffic simulation integration:} End-to-end coupling with real-time traffic simulation (SUMO) for flood-aware routing, where the 15 min forecast horizon provides advance warning for proactive road closures.
    \item \hl{Generalization:} Extension to other coastal urban areas beyond Norfolk to evaluate the transferability of the physics-informed loss framework.
\end{enumerate}

\vspace{6pt}

\authorcontributions{Conceptualization, L.D.\ and R.C.; methodology, L.D.; software, L.D.; validation, L.D., Y.W.\ and J.L.G.; formal analysis, L.D.; investigation, L.D.; resources, Y.W.\ and J.L.G.; data curation, Y.W.; writing, original draft preparation, L.D.; writing, review and editing, L.D., Y.W., J.L.G.\ and R.C.; visualization, L.D.; supervision, R.C.\ and J.L.G.; project administration, R.C.; funding acquisition, R.C. All authors have read and agreed to the published version of the manuscript.}

\funding{\hl{This research} %
 was funded by the Commonwealth Cyber Initiative (CCI) Coastal Virginia Node (CVN).}

 \dataavailability{\hl{The original contributions presented in this study are included in the article. Further inquiries can be directed to the corresponding authors.} %
 }

\acknowledgments{\hl{The authors} %
 acknowledge the Chandra Robot Autonomy Lab and the Hydroinformatics Research Group at the University of Virginia for supporting this work, and the \mbox{\hl{Wang et al.}~\cite{wang2026hybrid}} %
 team for providing access to the Norfolk flood dataset.}

\conflictsofinterest{\hl{The authors declare no conflicts of interest.} %
}

\abbreviations{Abbreviations}{
The following abbreviations are used in this manuscript:
\\

\noindent 
\begin{tabular}{@{}ll}
ASPP    & Atrous Spatial Pyramid Pooling\\
CNN     & Convolutional Neural Network\\
DEM     & Digital Elevation Model\\
LSTM    & Long Short-Term Memory\\
MAE     & Mean Absolute Error\\
MSE     & Mean Squared Error\\
PINN    & Physics-Informed Neural Network\\
RMSE    & Root Mean Squared Error\\
ROM     & Reduced-Order Model\\
SWE     & Shallow Water Equations\\
TWI     & Topographic Wetness Index\\
WSE     & Water Surface Elevation
\end{tabular}
}

\begin{adjustwidth}{-\extralength}{0cm}

\reftitle{References}

\PublishersNote{}
\end{adjustwidth}

\end{document}